\documentclass{article}
\usepackage{spconf,amsmath,graphicx}
\usepackage[pagebackref=true,breaklinks=true,letterpaper=true,colorlinks,bookmarks=false]{hyperref}

\title{The Mind's Eye: Visualizing Class-Agnostic Features of CNNs}
\name{Alexandros Stergiou \thanks{Thanks to the Netherlands Organization for Scientific Research (NWO) for funding this research with a TOP-C2 grant (ARBITER).}}
\address{
Department of Information and Computing Sciences, Utrecht University, Utrecht, Netherlands\\
\normalsize a.g.stergiou@uu.nl}
\begin{document}
%
\maketitle
\begin{abstract}
Visual interpretability of Convolutional Neural Networks (CNNs) has gained significant popularity because of the great challenges that CNN complexity imposes to understanding their inner workings. Although many techniques have been proposed to visualize class features of CNNs, most of them do not provide a correspondence between inputs and the extracted features in specific layers. This prevents the discovery of stimuli that each layer responds better to. We propose an approach to visually interpret CNN features given a set of images by creating corresponding images that depict the most informative features of a specific layer. Exploring features in this class-agnostic manner allows for a greater focus on the feature extractor of CNNs. Our method uses a dual-objective activation maximization and distance minimization loss, without requiring a generator network nor modifications to the original model. This limits the number of FLOPs to that of the original network. We demonstrate the visualization quality on widely-used architectures.\footnote{Code is available at \url{https://git.io/JL9Wg}\\ and our demo video: \url{https://youtu.be/Au3jaUdnPKM}} 

\end{abstract}
\section{Introduction}
\label{sec:intro}

\begin{figure}[ht]
\includegraphics[width=\linewidth]{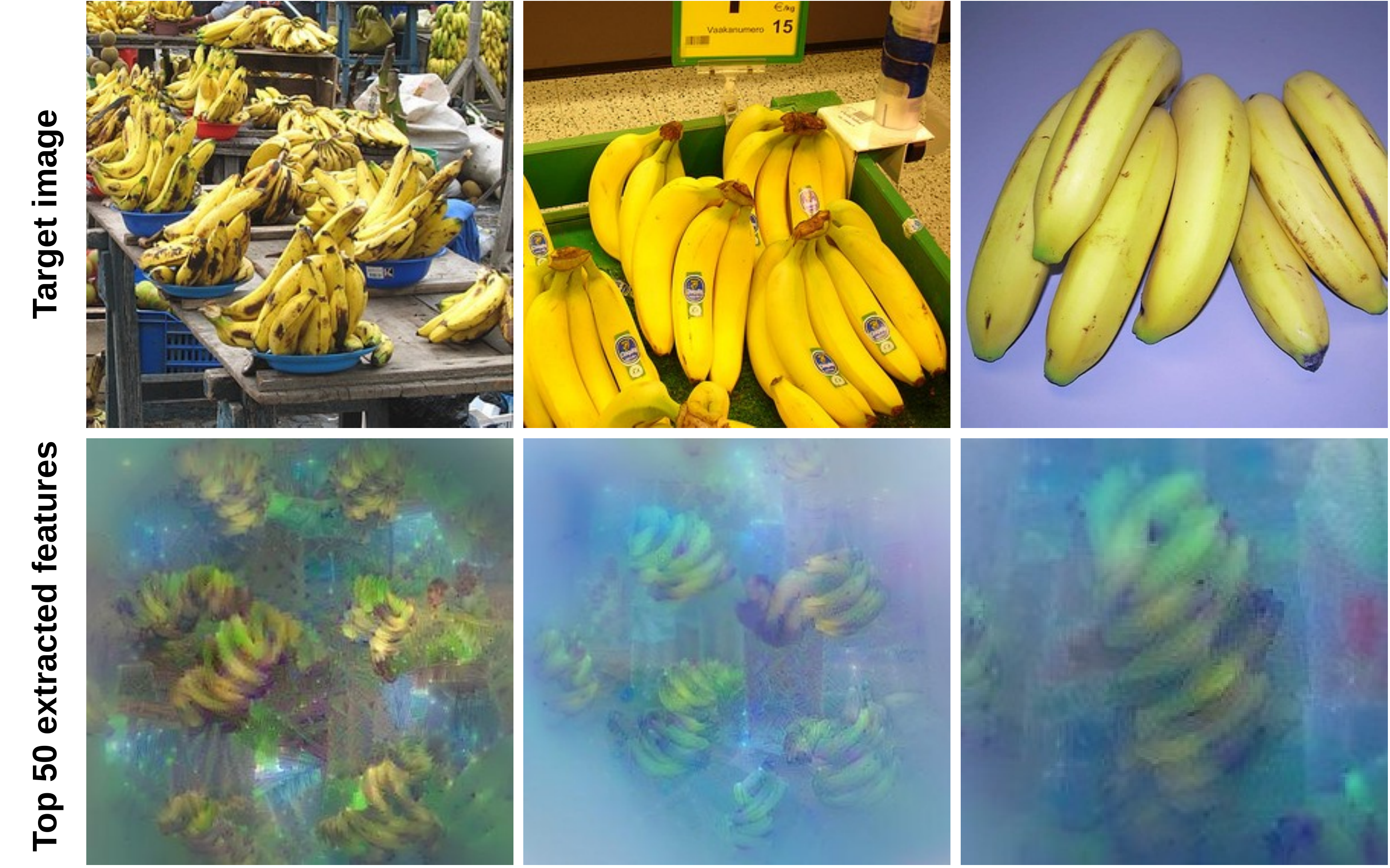}
\caption{\textbf{Top 50 extracted features}. ResNet-50 \cite{he2016deep} was used with features visualized from `layer6.5.conv1'.}
\label{fig:target_feats}
\vspace{-3mm}
\end{figure}

Deep learning architectures have achieved substantial breakthroughs in comparison to hand-coded feature extractors, for a wide variety of image and video tasks. While their performance is high, their interpretability remains limited.

For this reason, methods on visualizing features of CNNs have received significant research interest over the last few years. We identify two main approaches. The first is to consider the image regions that networks find informative \cite{fong2017interpretable,montavon2017explaining,selvaraju2017grad}. This approach allows for the selection of salient regions, but it does not provide a description for the feature's appearance. The second set of methods addresses this shortcoming by explicitly creating visualizations that activate features of specific classes \cite{erhan2009visualizing,zeiler2014visualizing}.

While both classes of approach have shown great promise to establish robust visual explanations for CNN features, one key aspect of deep learning method has not been explicitly addressed. Although early extracted features can be easily interpreted based on edges and textural patterns, features of deeper layers are significantly more complicated while they mostly do not correspond to singular concepts. We refer to features with polysemantic interpretations as \textit{entangled} \cite{mu2020compositional}.

To visualize neurons that encapsulate entangled features, we propose a multi-objective method that creates class-agnostic visual representations of image features. As a result, we can show the top $k$ descriptive features in sets of images. Our approach further addresses the common problem of different interpretations through perturbations \cite{ghorbani2019interpretation}, as the method is not constrained by visualizing features in a per-class fashion. An example appears in Figure~\ref{fig:target_feats}.

Our contributions are as follows:

\begin{itemize}
    \item We propose a class-agnostic method for visual explanations of CNN features. Our method uses a multi-objective loss based on activation maximization that optimizes an input image through the excitation of a user-defined number of layer neurons.
    \item We design an axiomatic distance objective to address \textit{entangled} features by minimizing the distance between produced image activations and the averaged target activations of real images.
    \item We test our approach on different layers and neurons of popular CNNs and show that our visualizations can uncover interesting features in sets of images.
\end{itemize}

The remainder of the paper is structured as follows. We first summarize related work on visual interpretations. We detail our method in Section~\ref{sec:methodology}. We demonstrate the produced feature visualizations in Section~\ref{sec:results} and conclude in Section~\ref{sec:conclusions}.

\section{Related Work}
\label{sec:related}

Recent works have argued the importance of interpretability in CNNs \cite{bau2017network,huk2018multimodal} and how it can further lead to CNN improvements \cite{hendricks2018grounding}. However, creating methods that capture the inner workings of CNNs has been proven challenging. 

One widely used method to visualize CNN features is to maximize neurons that correspond to a specific class \cite{erhan2009visualizing}. To maximize a class neuron, the input is composed of trainable parameters that are updated based on the gradients. As the sole consideration of class activations does not give an intuitive representation of the features that correspond to classes, Zeiler and Fergus \cite{zeiler2014visualizing} proposed a de-convolution approach that aims at approximating layer features. Later works of Simonyan \textit{et al.} \cite{simonyan2013deep} and Nguyen \textit{et al.} \cite{nguyen2015deep} have shown how the exclusive use of an activation maximization objective can lead to the creation of unrealistic images, since the space of possible images and patterns that can be produced and that are close to extracted patterns, is extremely vast. This motivated the exploration of regularization techniques aimed at constraining the space of possible visual representations. Some of these techniques include the use of Gaussian filters \cite{yosinski2015understanding,wang2018visualizing} during the image optimization process, jitter effect \cite{mordvintsev2015inceptionism}, and creating center-biased gradient masks \cite{nguyen2016multifaceted}. 

Other approaches to improve the realism in images consider using a separate network that is capable of synthesizing feature visualizations \cite{baumgartner2018visual,nguyen2016synthesizing} based on adversarial training similar to that of generative networks. Although the visual quality in generative models is higher, generators lack in terms of representing the causality of learned features \cite{holzinger2019causability}, as they include an additional factor of ambiguity through the generator sub-network. 

To address the problems associated with activation-maximization, we propose a method that is inspired by recent distance-minimization-based generative networks. Our method optimizes an image-based dot product of activations from generated and real-world images while additionally decreasing their activation distances within the feature space.

\begin{figure}[ht]
\begin{center}
\includegraphics[width=0.9\linewidth]{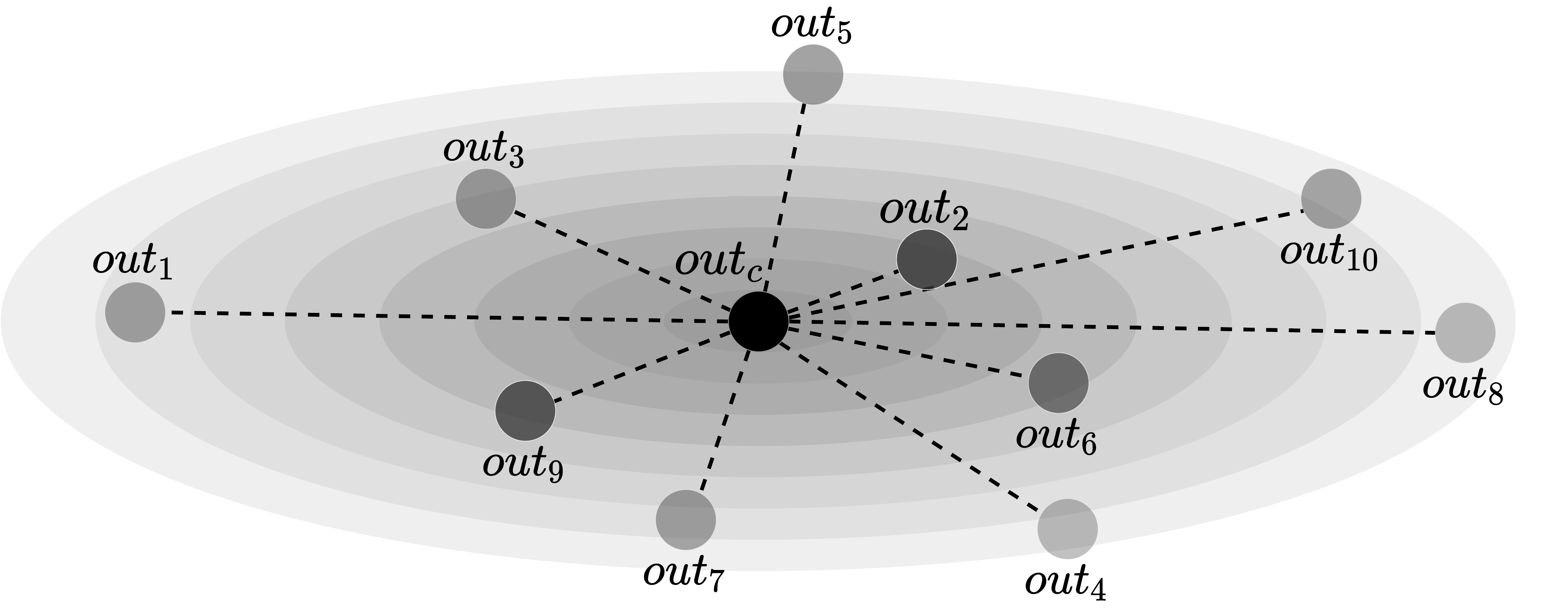}
\caption{\textbf{Feature space distances}. Point corresponds to top-10 embeddings ($out_{n},\; n \in N$) with center $out_{c}$.}
\end{center}
\label{fig:distance}
\vspace{-2mm}
\end{figure}

\section{Visualization methodology}
\label{sec:methodology}

We use a dot-product activation maximization with an additional distance minimization regression objective to optimize a trainable input image to represent the most informative features for a specific layer in the model. 

\subsection{Multi-faced clustered neuron selection}
\label{ssec:clustering}

We use a multi-faceted technique similar to the one proposed by Nguyen \textit{et al.} \cite{nguyen2016multifaceted} to optimize the creation of the initial images ($\overline{img}$) as well as the target activations ($t_{i}$) of layer $i$ used for regressions. We define $C$ target facets. We use real images and perform a normal forward-pass in the network until layer $i$. We then reduce the original channel dimensionality of activations $a_{i}$ in layer $i$ through PCA \cite{jolliffe2003principal} and t-SNE \cite{maaten2008visualizing} to create 2-dimensional embeddings $out$ that are then clustered into $C$ clusters with \textit{k-Means} \cite{lloyd1982least}. Instead of using the average within each of the $C$ clusters $c$ ($1 \leq c \leq C$), we consider the $N = 10$ 2D embeddings closest to each cluster center ($out_{c}$) to allow for a better correspondence to feature activations that are characteristic for cluster $c$. Based on the euclidean distance between the 2D embeddings and the cluster center ($edist(out_{c},out_{n}) \forall n \in N$) we create a weight penalty $w_{c,n}$. The weight corresponds to the effect of each activation ($a_{i,n}$) that is exponentially counter-equal to the euclidean distance between its 2D embedding $out_{n}$ and the cluster center $out_{c}$. This is summarized in Eq.~\ref{eq:softdist} with the image ($\overline{img}_{c}$) initialization and the discovered target activations ($t_{i,c}$). We include a constant ($\gamma = 1e^{-5}$) for numeric stability.
\vspace{-1.5mm}

\begin{equation}
\label{eq:softdist}
\begin{split}
{w}_{c,n} = \frac{e^{edist(out_{c},out_{n})^{-1} + \gamma}}{\sum \limits_{n \in N} e^{edist(out_{c},out_{n})^{-1} + \gamma}} \qquad \qquad \\
\overline{img}_{c} = \sum \limits_{n \in N} img_{n} * w_{c,n} \; \; \& \; \; t_{i,c} = \sum \limits_{n \in N} a_{i,n} * w_{c,n}
\end{split}
\vspace{-1.5mm}
\end{equation}

The effect is visualized in Figure~\ref{fig:distance}, where the distance between each 2D embedding $out_{n}$ and the corresponding cluster center $out_{c}$ determines the contribution of each activation $a_{i,n}$ to the final target activation $t_{i,c}$.

\begin{figure*}[t!]
\includegraphics[width=\textwidth]{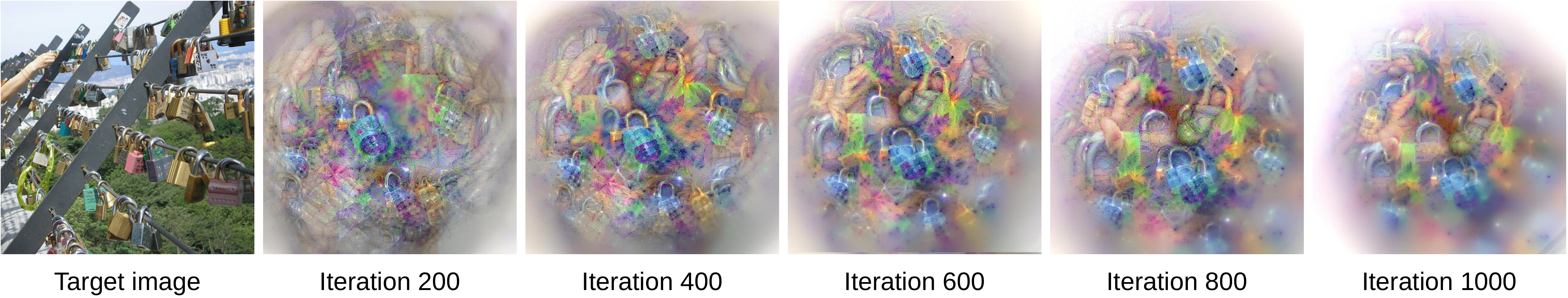}
\caption{\textbf{Image optimization iterations}. The target image from ImageNet's \cite{ILSVRC15} padlock class is the image closest to the cluster center. The top 30 features of wide-ResNet101's \cite{zagoruyko2016wide} `layer6.12.conv3' are used.}
\label{fig:iterations}
\vspace{-2mm}
\end{figure*}

\subsection{Objective formalization}
\label{ssec:objectives}

We define our loss function based on two additional auxiliary objectives to improve the feature clarity while simultaneously reduce the effects of \textit{feature entanglement}. To visualize specific features, we select the top $k$ features of each target activation $t_{i,c}$ of each cluster $c$. This creates an averaged overview over the most informative top $k$ features that should be visualized.

We define a \textit{dot-product activation maximization loss} ($DM$) as the channel-wise dot product of the produced activation maps $a_{i}$ and the discovered target layer activations $t_{i,c}$. This calculation is performed for all top $k$ channels:

\begin{equation}
\label{eq:activation_maximization}
DM = \sum\limits_{j \in k} a_{i,j} * t_{i,c,j}
\end{equation}

Through the maximization of the dot-product of the produced activations and the target layer activations, we create a path towards meaningful features for the gradients. However, this also corresponds to larger \textit{feature entanglement} as the span of possible gradient directions to provide positive improvements can be extensive.

To address this issue, we include a second objective: a multi-dimensional distance minimization between the produced activation maps $a_{i}$ and the target activations $t_{i}$. For the distance loss function, we use a super-set of distance methods as proposed by Barron \cite{barron2019general} (denoted as $AD$). Because of the heteroscedastic nature of the produced distances, using trainable parameters ($r,b$), shown in Eq.~\ref{eq:dist}, can better fit the produced multivariate distances.
\vspace{-1.5mm}

\begin{equation}
\label{eq:dist}
\begin{split}
AD = \begin{cases}
\frac{1}{2} (\frac{mdist}{r})^{2}, \; for \; b = 2 \\
log(\frac{1}{2}(\frac{mdist}{r})^{2}+1) \; for \; b = 0 \\
1-exp(-\frac{1}{2}(\frac{mdist}{r})^{2}) \; for \; b \rightarrow -\infty \\
\frac{|b-2|}{b} \left( \left( \frac{(\frac{mdist}{r})^{2}}{|b-2|}+1 \right)^{b/2}+1 \right) \; otherwise 
\end{cases}\\
where, \: \:mdist = \left( \sum\limits_{j \in k}||a_{i,j} - t_{i,c,j}|| \right) \qquad \quad 
\end{split}
\end{equation}

Finally, we synthesize our loss function $\mathcal{L}$ from Eqs.~\ref{eq:activation_maximization} and \ref{eq:dist} by including a penalty for the activations of the previous layers ($a_{i-1}$) with channel size $D$, through $L_{1}$ regularization. The scaling value $\lambda$ has an initial value of $1e^{-3}$ which is linearly decreased to $1e^{-4}$ during training. The final loss is:
\vspace{-1.5mm}

\begin{equation}
\label{eq:Loss}
\mathcal{L} = AD - DM + \lambda * \sum \limits_{j \in D} ||a_{i-1,j}||
\vspace{-1.5mm}
\end{equation}

\subsection{Parameterization setup}
\label{ssec:setup}

We include parameterized noise functions within our workflow as constraints for the high-frequency gradients during back-propagation. This allows for the minimization of possible feature imbalances \cite{wang2018visualizing} and improves the final visual representation quality of CNN features. We include popular techniques used in visualization methods such as image blurring with low-variance Gaussian kernels \cite{yosinski2015understanding,wang2018visualizing} in combination with a denoising split Bregman algorithm \cite{getreuer2012rudin}. We additionally include a center-based gradient mask \cite{nguyen2016multifaceted} to limit feature cluttering as well as to limit feature duplication during training. Learning rate $lr$, standard deviations of the Gaussian blur, denoising, and center-based gradient mask ($\sigma$) are decreased linearly over each iteration $t$ ($1 \leq t \leq T$), based on user-set starting ($start$) and final ($end$) values:

\begin{equation}
\label{eq:variable_update}
lr,\sigma = \frac{((end-start)*t)+(start*T)}{T-1}
\vspace{-1.5mm}
\end{equation}

We present iterative changes in Figure~\ref{fig:iterations} where an image is optimized to visualize features that correspond to a target.

\begin{figure*}[t!]
\includegraphics[width=\textwidth]{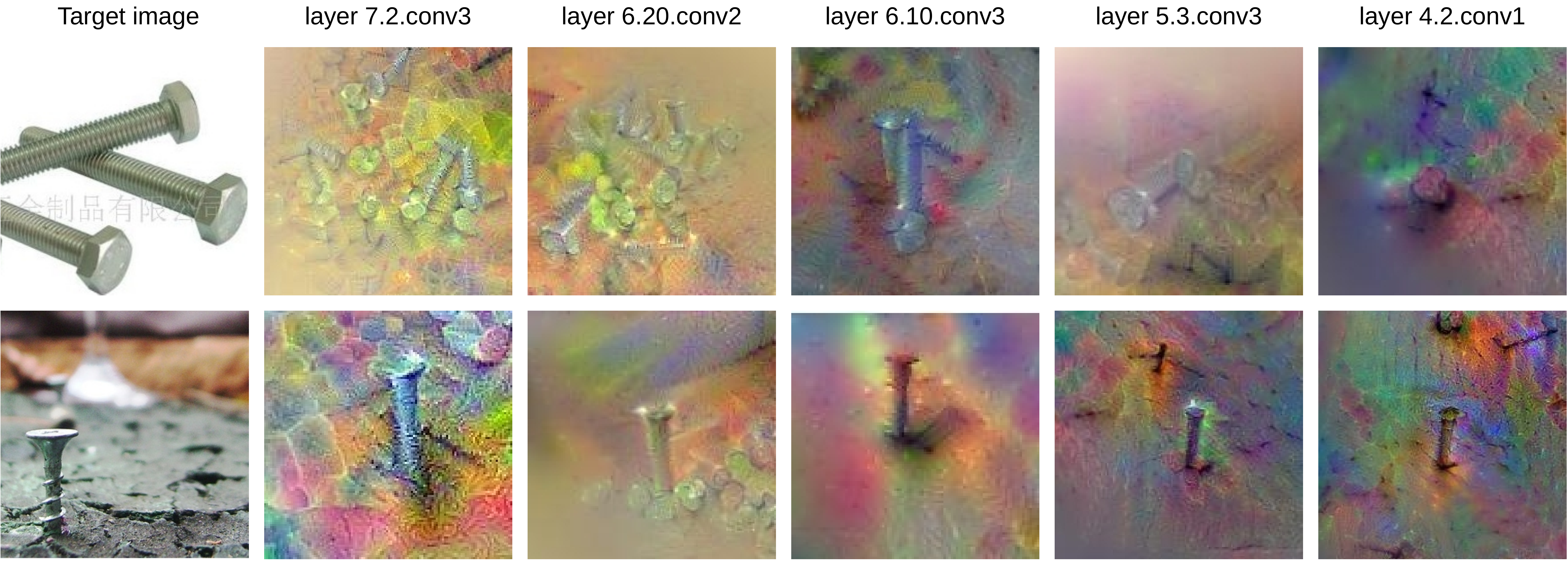}
\caption{\textbf{Cross-layer visualizations of top-100 features}. Images were sampled from ImageNet's `screw' class. The targets include variations in orientation, number of objects and shape. We use ResNeXt-101 \cite{xie2017aggregated} with the corresponding optimization layers being displayed at the top of each column.}
\label{fig:examples}
\vspace{-2mm}
\end{figure*}

\section{Visualizations}
\label{sec:results}

We demonstrate in Figure~\ref{fig:target_feats} three cases of the same object where the produced feature visualizations present some degree of variation. Although the target images show the same object (from class `banana' in ImageNet), the number of objects present may correspond to differences in terms of the most dominant feature activations. We note however, that differences in the image targets do not present detrimental effects in the overall feature combination.

In Figure~\ref{fig:iterations}, we present how features are visualized over time. This shows the variations in regions of the image being optimized and how feature-inclusive regions change over time. This demonstrates the ability of a specific architecture to combine only a certain number of features in order to encapsulate the general appearance of an object. For example, for images from the `padlock' class, the 30 most highly activated features can be seen as sufficient for describing the overall appearance of the object. This can aid in determining the images and classes that are easier or harder for the network to extract meaningful features from.

Lastly, we show in Figure~\ref{fig:examples} the features that are extracted among different layers of the same network. Using images from ImageNet’s `screw' class, we sample target sets of images varying in terms of their general orientation with objects being perpendicular or at an angle, as well as differences in screw head types and number of objects. Visualizations of later layers (7.2 and 6.20) include the metallic look of the object as well as screw body patterns and detail in terms of the screw head type (circular and hexagonal). Such specific features however are absent in earlier layers as their feature extractors seem to correspond to basic features in terms of the general object appearance with the number of information-rich \textit{entangled} features being lower from that of later layers.

\section{Conclusions}
\label{sec:conclusions}

We have proposed a novel feature visualization method aimed at providing visual explanations for the top features extracted by CNN layers. Images are optimized based on a dual-objective loss which includes the dot-product activation maximization and the distance minimization between the produced and target layer activations. To reduce feature overlap and to improve the overall visualization clarity, we apply blurring and de-blurring filters as well as a gradient mask to the generated images during optimization.

We present corresponding CNN features that are associated with specific classes and images. Based on this, we believe that the produced visual explanations can improve the in-depth understanding of trained CNNs and the features with polysemantic interpretations that are associated with a specific image or set of images.



\bibliographystyle{IEEEbib}
\bibliography{refs}

\end{document}